%% file: paper.tex
\documentclass[10pt,letterpaper]{article}
\usepackage{spconf}
\usepackage[T1]{fontenc}
\usepackage{amsmath,amssymb}
\usepackage{graphicx,booktabs,array,multirow}
\usepackage{cite,microtype,url}
\usepackage[hidelinks]{hyperref}
\usepackage{balance}

\newcommand{\method}{\textsc{SampleSelect}}
\newcommand{\fullrep}{\textsc{FullRep}}
\newcommand{\sg}{\operatorname{stopgrad}}
\newcommand{\CE}{\operatorname{CE}}
\newcommand{\TopK}{\operatorname{TopK}}
\newcommand{\RelaxedTopK}{\operatorname{RelaxedTopK}}
\DeclareMathOperator{\GAP}{GAP}
\DeclareMathOperator{\vecop}{vec}

\newenvironment{methodequation}{%
  \begingroup
  \small
  \setlength{\jot}{1pt}%
  \setlength{\abovedisplayskip}{3.2pt plus .8pt minus 1.2pt}%
  \setlength{\belowdisplayskip}{3.2pt plus .8pt minus 1.2pt}%
  \setlength{\abovedisplayshortskip}{2pt plus .5pt}%
  \setlength{\belowdisplayshortskip}{2.5pt plus .5pt minus .5pt}%
  \begin{equation}%
}{%
  \end{equation}%
  \endgroup
}

\title{Sample-Conditioned Representation Selection\newline for Audio Few-Shot Learning}
\name{{\small
Fengrui Liu$^{1,2,*,\ddagger,\mathsection}$,
Ningxin Shen$^{3,*}$,
Yi Li$^{1}$,
Yiwei Fu$^{2}$,
Feng Liu$^{4}$, \textit{Senior Member, IEEE},
Jiangmeng Li$^{1},\dagger$
}}

\address{
\parbox{0.88\textwidth}{\centering
$^{1}$National Key Laboratory of Space Integrated Information System,
Institute of Software, Chinese Academy of Sciences\\
$^{2}$School of Computer Science and Technology,
East China Normal University\\
$^{3}$School of Computer Science,
Nanjing University\\
$^{4}$School of Psychology,
Shanghai Jiao Tong University\\
$^{*}$Equal contribution. \quad
$^{\ddagger}$Project lead. \quad
$^{\dagger}$Corresponding author. \quad
$^{\mathsection}$Work done while the first author was an intern at the
Institute of Software, Chinese Academy of Sciences.
}
}
\begin{document}
\ninept
\maketitle
\input{sections/00_abstract}
\input{sections/01_introduction}
\input{sections/03_method}
\input{sections/04_experiments}

\input{sections/05_conclusion}
\clearpage
\balance
\bibliographystyle{IEEEtran}
\bibliography{references}
\end{document}

%% file: sections/00_abstract.tex
\begin{abstract}
Few-shot audio classifiers may rely on foreground--background co-occurrences and fail when those correlations shift. On SpurAudio, the resulting representation shift is concentrated and class dependent: for ResNet12, the top 10\% of channels explain 82.80\% of the null-corrected shift contribution. We propose \method, which predicts a fixed-budget feature mask independently for each input while keeping the encoder and source classifier frozen. Training uses differentiable Gumbel Top-$k$ selection with foreground classification and cross-background contrastive losses; inference uses deterministic Top-$k$ masks and support-only linear adaptation. Across ResNet12 and Conv64 in 5-way 1-shot and 5-shot evaluation, \method\ gives the best OOD accuracy among the compared methods and improves the matched full-representation control by 4.90--8.38 percentage points. Ablations and representation analyses further support the learned selection mechanism.Codes available at https://github.com/Cross-Innovation-Lab/SAMPLESELECT/
\end{abstract}


%% file: sections/01_introduction.tex
\section{Introduction}
\label{sec:intro}

Few-shot audio classification recognizes novel sound classes from limited labeled examples, commonly through episodic adaptation or source-trained representations~\cite{vinyals2016matching,snell2017prototypical,chen2019closer,liu2026physics,liu2026prp}. Real recordings also contain recurring foreground--background co-occurrences that models can exploit as shortcuts~\cite{geirhos2020shortcut}. When those associations change, previously useful features can become unreliable, while the small support set offers little evidence for identifying which coordinates remain trustworthy.

MetaAudio studies transfer across acoustic domains~\cite{heggan2022metaaudio}, whereas SpurAudio changes foreground--background co-occurrences between IID and OOD episodes~\cite{abuayoub2026spuraudio}. Existing representation adaptation and feature reweighting methods improve few-shot transfer~\cite{li2019categorytraversal,dvornik2020sur,li2022taskspecific,lee2022tdm,hu2018senet}, but do not directly target input-varying background sensitivity. This motivates selecting features for each example rather than always exposing the full representation or one global subset.

\input{sections/figure_overview}

A frozen ResNet12 reveals that the shift is highly concentrated and class dependent: the top 10\% of its 640 channels account for 82.80\% of the corrected shift contribution, and the affected channels vary substantially across foreground classes. We therefore propose \method, a fixed-budget, sample-conditioned selector. A lightweight scorer predicts feature importance from each input while the encoder and source classifier remain frozen. Training uses a differentiable Gumbel relaxation with foreground classification and grouped cross-background contrastive supervision~\cite{jang2017gumbelsoftmax,kool2019gumbeltopk,balin2019concrete,khosla2020supervised,sgouropoulos2025prototypicalcontrastive}; inference uses deterministic Top-$k$ masking and a support-only linear head.
\input{sections/figure_method}
On SpurAudio, across ResNet12 and Conv64 with 5-way 1-shot and 5-shot episodes, \method\ attains the highest OOD accuracy among the compared methods in all four settings and improves the matched \fullrep\ control by 4.90--8.38 percentage points. Ablations and further analyses support learned selection, cross-background supervision, and the functional relevance of the selected channels. Our contributions are: (1) evidence that background co-occurrence shift is concentrated and class dependent in frozen audio representations; (2) an inductive sample-conditioned fixed-budget selector trained without changing the source representation; and (3) consistent OOD gains across two backbones and two shot settings, supported by ablation and representation analyses.

%% file: sections/figure_overview.tex
\begin{figure}[t]
\centering
\includegraphics[width=0.9\columnwidth]{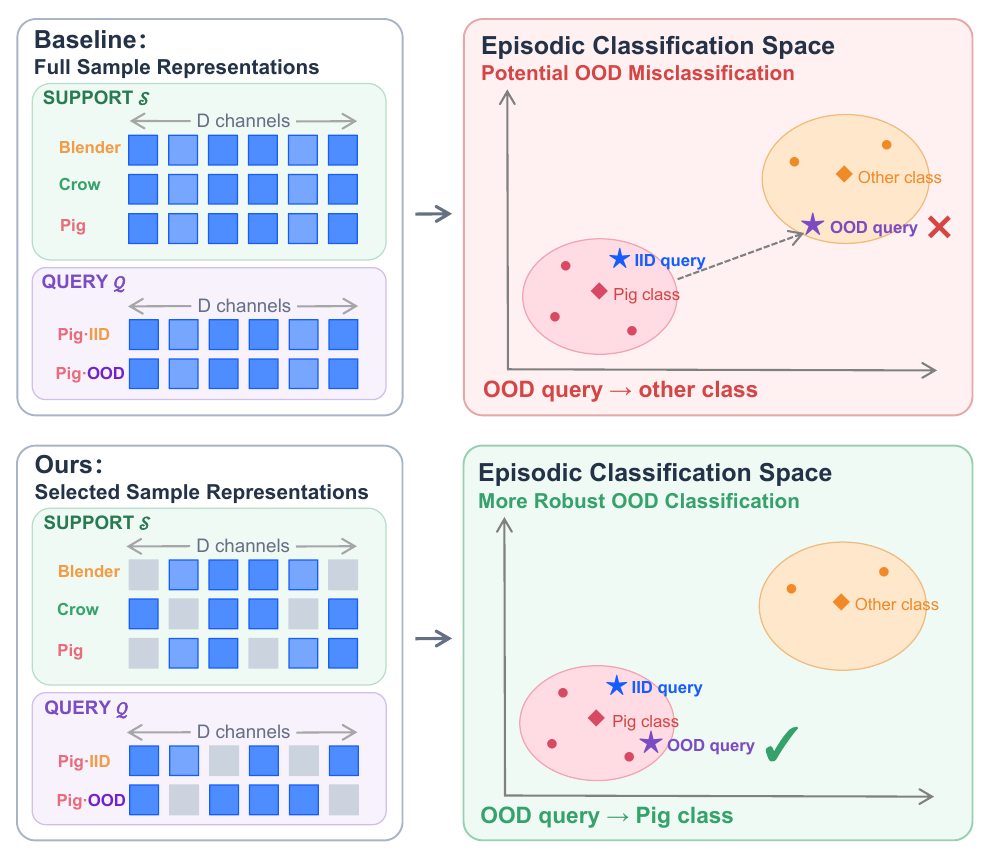}
\caption{Overview of few-shot classification under background shift. \fullrep\ passes all representation coordinates to a fresh episodic linear head; \method\ independently masks each support and query example before fitting and prediction. The class-space plots are schematic, not measured embeddings.}
\label{fig:overview}
\end{figure}

%% file: sections/figure_method.tex
\begin{figure*}[h]
\centering
\includegraphics[width=0.9\textwidth]{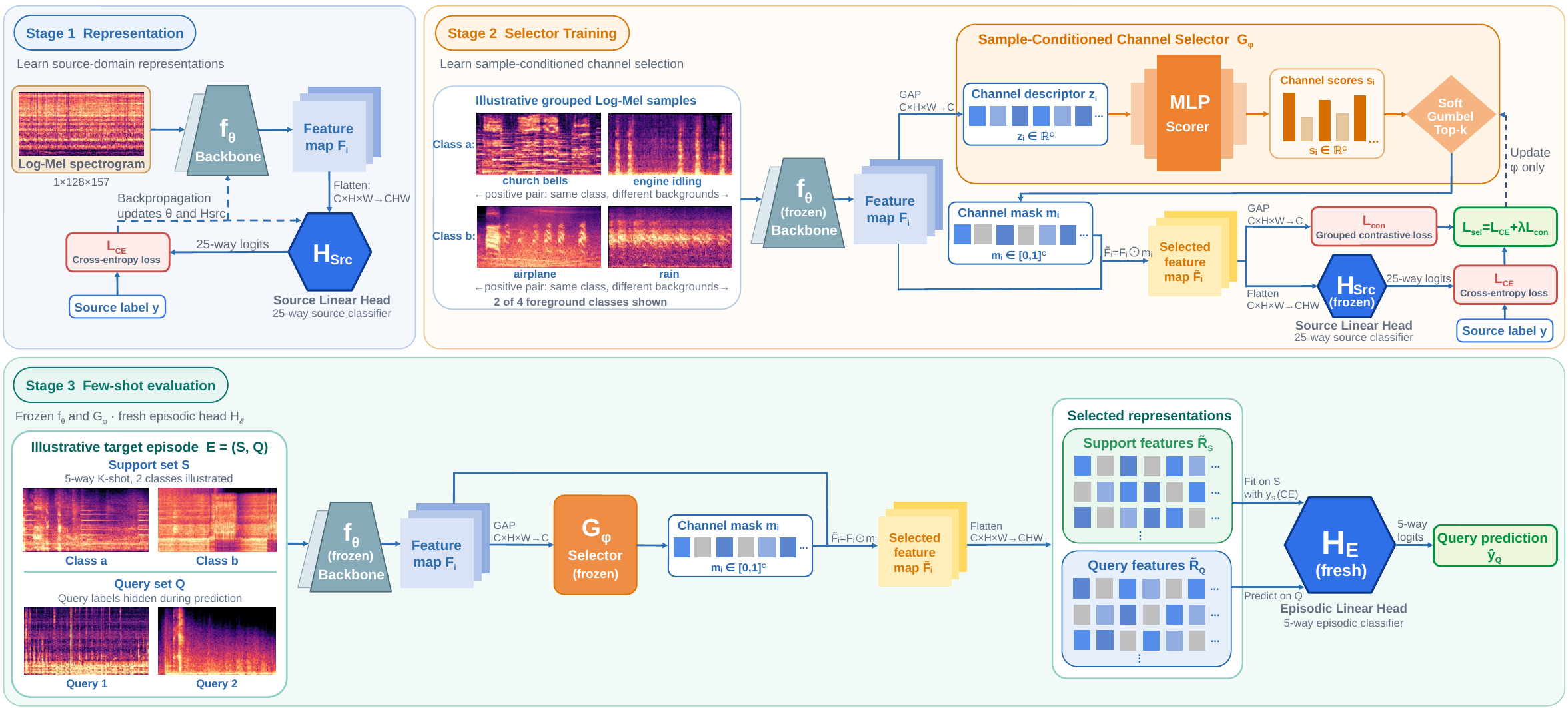}
\caption{Three-stage \method\ pipeline. The source encoder/head are trained then frozen; the selector learns sample-wise masks with classification and cross-background contrastive losses; inference uses deterministic Top-$k$ masks and a fresh support-only episodic head.}
\label{fig:method}
\end{figure*}

%% file: sections/03_method.tex
\section{methodology}
\label{sec:method}

Figure~\ref{fig:method} summarizes \method. We first learn a source representation, then freeze it and train only a sample-conditioned selector; few-shot evaluation uses deterministic masks and a fresh support-only head.
\begin{table*}[t]
\centering
\caption{Results on SpurAudio with ResNet12 and Conv64. Accuracy values are percentages.
$\Delta=\mathrm{IID}-\mathrm{OOD}$. \fullrep\ denotes the matched no-selection control.
Bold indicates the best IID or OOD accuracy among the compared methods.}
\label{tab:main_results}
\small
\setlength{\tabcolsep}{2.6pt}
\begin{tabular}{lcccccccccccc}
\toprule
& \multicolumn{6}{c}{ResNet12}
& \multicolumn{6}{c}{Conv64} \\
\cmidrule(lr){2-7}\cmidrule(lr){8-13}

& \multicolumn{3}{c}{1-shot}
& \multicolumn{3}{c}{5-shot}
& \multicolumn{3}{c}{1-shot}
& \multicolumn{3}{c}{5-shot} \\
\cmidrule(lr){2-4}
\cmidrule(lr){5-7}
\cmidrule(lr){8-10}
\cmidrule(lr){11-13}

Method
& IID & OOD & $\Delta$
& IID & OOD & $\Delta$
& IID & OOD & $\Delta$
& IID & OOD & $\Delta$ \\
\midrule

Baseline++ (2019)
& 57.694 & 53.078 & 4.616
& 75.032 & 65.490 & 9.542
& 50.705 & 46.667 & 4.038
& 65.286 & 56.934 & 8.352 \\

R2D2 (2019)
& 57.995 & 54.228 & 3.767
& 74.760 & 66.149 & 8.611
& 41.093 & 39.420 & 1.673
& 68.695 & 61.873 & 6.822 \\

ANIL (2020)
& 54.129 & 50.893 & 3.236
& 66.015 & 58.352 & 7.663
& 49.522 & 46.499 & 3.023
& 64.763 & 56.009 & 8.754 \\

BDCSN (2022)
& 61.241 & 57.961 & 3.280
& 73.564 & 66.691 & 6.873
& 44.992 & 42.374 & 2.618
& 58.951 & 55.031 & 3.920 \\

PADDLE (2022)
& 52.196 & 46.842 & 5.354
& 70.380 & 59.299 & 11.081
& 50.016 & 44.267 & 5.749
& 66.767 & 55.912 & 10.855 \\

Proto-LP (2023)
& 59.852 & 56.036 & 3.816
& 74.762 & 67.454 & 7.308
& 57.054 & 53.446 & 3.608
& 69.916 & 61.328 & 8.588 \\

BPA (2024)
& 60.325 & 55.828 & 4.497
& 75.964 & 63.516 & 12.448
& 49.158 & 45.369 & 3.789
& 66.173 & 50.924 & 15.249 \\

ECPE (2026)
& 61.397 & 56.391 & 5.006
& 74.890 & 65.642 & 9.248
& 55.836 & 51.974 & 3.862
& 69.038 & 59.536 & 9.502 \\

\midrule

\fullrep
& 57.169 & 51.997 & 5.172
& 74.369 & 63.259 & 11.110
& 51.470 & 47.032 & 4.438
& 67.641 & 56.813 & 10.828 \\

\method
& \textbf{62.307} & \textbf{57.979} & 4.327
& \textbf{77.523} & \textbf{68.161} & 9.362
& \textbf{57.357} & \textbf{53.635} & 3.721
& \textbf{72.993} & \textbf{65.188} & 7.805 \\

\bottomrule
\end{tabular}
\end{table*}
\subsection{Source Representation Learning}
For input $x_i$ with foreground label $y_i$, the selectable descriptor $z_i$ and classifier representation $h_i$ are
\begin{methodequation}
\begin{aligned}
\text{ResNet12: }&F_i=f_\theta(x_i)\in\mathbb{R}^{640\times4\times5},\quad z_i=\GAP(F_i),\\
&h_i=\vecop(F_i)\in\mathbb{R}^{12800},\\
\text{Conv64: }&z_i=h_i=f_\theta(x_i)\in\mathbb{R}^{1600}.
\end{aligned}
\label{eq:source_repr}
\end{methodequation}
We train the encoder and source classifier $H_\psi$ on source foreground classes,
\begin{methodequation}
(\theta^\star,\psi^\star)=\arg\min_{\theta,\psi}\frac1N\sum_{i=1}^N \CE\!\left(H_\psi(h_i),y_i\right),
\label{eq:source_training}
\end{methodequation}
and freeze both thereafter.

\subsection{Sample-Conditioned Selector Learning}
A scorer predicts one score per selectable coordinate, with retention ratio $r$:
\begin{methodequation}
\begin{aligned}
s_i&=G_\phi\!\left(\sg(z_i)\right)\in\mathbb{R}^{D},\qquad k=\lfloor rD\rfloor,\\
m_i^{\mathrm{tr}}&=\RelaxedTopK_{\tau}(s_i,k)\in[0,1]^D,
\end{aligned}
\label{eq:selector_mask_train}
\end{methodequation}
where $D=640$ for ResNet12 and $1600$ for Conv64. $G_\phi$ acts independently on each input. During training, sequential Gumbel-Softmax draws without replacement provide a differentiable Top-$k$ relaxation~\cite{jang2017gumbelsoftmax,kool2019gumbeltopk}; exact cardinality is enforced at inference.

The selected classifier representation is
\begin{methodequation}
\widetilde h_i=
\begin{cases}
\vecop\!\left(m_i\odot_c F_i\right), & \text{ResNet12},\\
m_i\odot h_i, & \text{Conv64},
\end{cases}
\label{eq:masked_repr}
\end{methodequation}
where $\odot_c$ broadcasts a channel mask over spatial locations. Zero masking preserves coordinate alignment across examples.

The frozen source head preserves foreground information through
\begin{methodequation}
\mathcal L_{\mathrm{cls}}=\frac1B\sum_{i=1}^{B}\CE\!\left(H_{\psi^\star}(\widetilde h_i),y_i\right).
\label{eq:selector_ce}
\end{methodequation}
For background group $b_i$, define $P(i)=\{p\ne i:y_p=y_i,\ b_p\ne b_i\}$. Let $q_i=\GAP(m_i\odot_cF_i)$ for ResNet12 and $q_i=\widetilde h_i$ for Conv64, with normalized similarity $u_{ij}=(q_i/\|q_i\|_2)^\top(q_j/\|q_j\|_2)$. For a valid anchor,
\begin{methodequation}
\mathcal L_{\mathrm{con}}^{(i)}=-\frac1{|P(i)|}\sum_{p\in P(i)}
\log\frac{\exp(u_{ip}/T)}{\sum_{a\in G(i)\setminus\{i\}}\exp(u_{ia}/T)}.
\label{eq:contrastive}
\end{methodequation}
where $G(i)$ is its comparison group. We average over anchors with valid positives and minimize $\mathcal L_{\mathrm{sel}}=\mathcal L_{\mathrm{cls}}+\lambda_{\mathrm{con}}\mathcal L_{\mathrm{con}}$, updating only $\phi$. Thus classification preserves category information, while contrastive supervision favors same-class consistency across observed backgrounds. Background labels are used only in selector training.

\subsection{Inductive Few-Shot Evaluation}
At inference, each example receives an independent deterministic mask
\begin{methodequation}
m_{i,c}^{\mathrm{eval}}=\mathbf 1\!\left[c\in\TopK(s_i,k)\right],\qquad \|m_i^{\mathrm{eval}}\|_0=k.
\label{eq:eval_mask}
\end{methodequation}
For a 5-way $K$-shot episode $\mathcal E=(\mathcal S,\mathcal Q)$, a fresh linear head is fitted only on masked support representations and then applied to each masked query:
\begin{methodequation}
\begin{aligned}
W_{\mathcal E}^{\star}&=\arg\min_W\frac1{|\mathcal S|}\sum_{(x_j,y_j)\in\mathcal S}\CE\!\left(W\widetilde h(x_j),y_j\right),\\
\widehat y_q&=\arg\max_c\left[W_{\mathcal E}^{\star}\widetilde h(x_q)\right]_c.
\end{aligned}
\label{eq:episode_eval}
\end{methodequation}
No query label, query-set statistic, or test-time background label is used for masking or adaptation. The matched \fullrep\ control uses the same support-only procedure with $m_i=\mathbf 1$.

%% file: sections/04_experiments.tex
\section{Experiments}
\label{sec:experiments}

\subsection{Experimental Setup}

\noindent\textbf{Datasets.}
We evaluate on SpurAudio~\cite{abuayoub2026spuraudio}, which contains 25 training, 5 validation, and 8 test foreground classes and controls foreground--background co-occurrence to construct IID and OOD few-shot episodes. We follow the standard 5-way 1-shot and 5-shot protocols with 10 query examples per class and report the mean over 1,000 episodes. We report IID accuracy, OOD accuracy, and the shift gap
$\Delta_{\mathrm{shift}}=\mathrm{Acc}_{\mathrm{IID}}-\mathrm{Acc}_{\mathrm{OOD}}$.
OOD accuracy is the primary metric.

\noindent\textbf{Backbones and Baselines.}
We evaluate two representation architectures. ResNet12 performs channel-level selection over 640 channels, while Conv64 performs coordinate-level selection over a 1,600-dimensional projected representation. We compare with representative methods reported under the SpurAudio protocol and include \fullrep\ as a matched control using the same frozen backbone and episodic classifier without representation selection, thereby isolating the effect of selection.

\noindent\textbf{Implementation Details.}
Inputs are standardized $1\times128\times157$ log-Mel spectrograms. The source encoder is trained for 30 epochs and then frozen. ResNet12 uses retention ratio $r=0.7$ with $k=448$, while Conv64 uses $r=0.8$ with $k=1280$. Unless otherwise stated, selector training uses Gumbel temperature $\tau=0.3$, contrastive temperature $T=0.07$, and $\lambda_{\mathrm{con}}=0.02$. During few-shot evaluation, the encoder and selector remain frozen and only a support-only linear classifier is optimized for each episode.
\subsection{Main Results}
\label{sec:main_results}

\begin{figure*}[t]
\centering
\begin{minipage}[t]{0.32\textwidth}
    \centering
    \includegraphics[width=\linewidth]{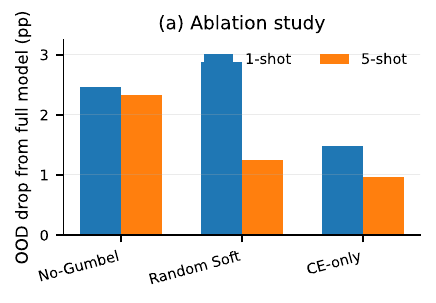}
\end{minipage}\hfill
\begin{minipage}[t]{0.32\textwidth}
    \centering
    \includegraphics[width=\linewidth]{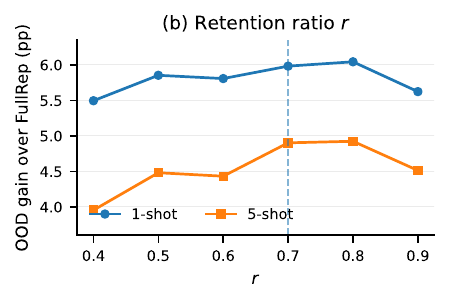}
\end{minipage}\hfill
\begin{minipage}[t]{0.32\textwidth}
    \centering
    \includegraphics[width=\linewidth]{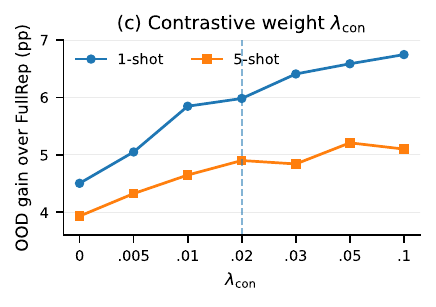}
\end{minipage}
\caption{Ablation and sensitivity analysis on ResNet12. (a) OOD accuracy drop after removing individual components of \method. (b) OOD gain over \fullrep\ for different retention ratios $r$. (c) OOD gain over \fullrep\ for different contrastive weights $\lambda_{\mathrm{con}}$. Dashed lines indicate the default settings used in the main experiments.}
\label{fig:ablation_sensitivity}
\end{figure*}

Table~\ref{tab:main_results} shows that \method\ achieves the highest OOD accuracy among the compared methods in all four backbone and shot settings. It reaches 57.979\% and 68.161\% with ResNet12, and 53.635\% and 65.188\% with Conv64 for 1-shot and 5-shot evaluation, respectively. IID accuracy is also highest in each setting. Compared with the matched \fullrep\ control, \method\ improves OOD accuracy by 5.982 and 4.902 percentage points for ResNet12, and by 6.603 and 8.375 points for Conv64. The gains are consistent across both representation architectures and shot settings, showing that sample-conditioned selection improves the use of frozen representations under class--background co-occurrence shift.

\subsection{Ablation and Sensitivity Analysis}
\label{sec:ablation_sensitivity}

Figure~\ref{fig:ablation_sensitivity}(a) evaluates the main components of \method. Replacing learned selection with Random Soft decreases OOD accuracy by 2.882 and 1.254 points in the 1-shot and 5-shot settings. Removing the Gumbel relaxation reduces accuracy by 2.458 and 2.322 points, while removing the cross-background contrastive objective gives drops of 1.477 and 0.968 points. A learnable global selector achieves 55.799$\pm$1.372 and 66.869$\pm$1.004 OOD accuracy in the 1-shot and 5-shot settings, respectively, trailing \method\ by 2.180 and 1.292 points. These results support learned, differentiable, sample-conditioned selection and cross-background supervision.

Figures~\ref{fig:ablation_sensitivity}(b,c) study the two main hyperparameters. Performance changes only modestly for $r\in[0.4,0.9]$, and values near the default $r=0.7$ give similar OOD gains. For $\lambda_{\mathrm{con}}$, every tested positive value improves the OOD mean over $\lambda_{\mathrm{con}}=0$ in both shot settings. We use $\lambda_{\mathrm{con}}=0.02$ as a shared setting rather than tuning it separately for each evaluation condition.

\subsection{Further Analysis}
\label{sec:further}

The previous experiments establish the performance benefit of \method. We next examine the representation behavior that motivates sample-conditioned selection.

\begin{figure*}[t]
\centering
\begin{minipage}[t]{0.31\textwidth}
    \centering
    \includegraphics[width=\linewidth]{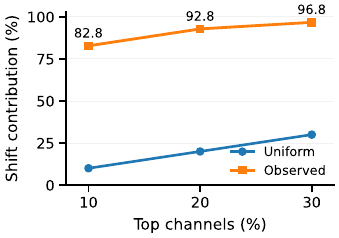}\\[-1mm]
    {\small (a) Shift concentration}
\end{minipage}\hfill
\begin{minipage}[t]{0.31\textwidth}
    \centering
    \includegraphics[width=\linewidth]{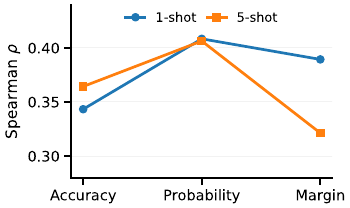}\\[-1mm]
    {\small (b) Shift and OOD degradation}
\end{minipage}\hfill
\begin{minipage}[t]{0.31\textwidth}
    \centering
    \includegraphics[width=\linewidth]{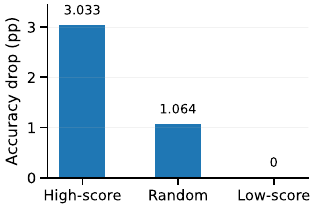}\\[-1mm]
    {\small (c) Functional blocking}
\end{minipage}
\caption{Further analysis on ResNet12. (a) Cumulative contribution of channels with the largest null-corrected representation shifts. (b) Mean Spearman correlation across three seeds between episode-level shift mass and IID-to-OOD degradation for accuracy, confidence, and classification margin. (c) Accuracy drop after blocking the highest-scored, random, or lowest-scored 30\% of channels while keeping the episodic classifier fixed.}
\label{fig:further_analysis}
\end{figure*}

\subsubsection{Concentrated and Class-Dependent Shift}
For foreground class $y$ and ResNet12 channel $c$, we measure the IID-to-OOD distribution change using the 1-Wasserstein distance and correct it with a within-class permutation null:
\begin{equation}
\begin{aligned}
d_{y,c}&=W_1\!\left(
\widehat P_{\mathrm{IID}}(z_c\mid y),
\widehat P_{\mathrm{OOD}}(z_c\mid y)
\right),\\[-1mm]
g_{y,c}&=d_{y,c}-\mathbb E_\pi[d^\pi_{y,c}],\qquad
\omega_{y,c}=
\frac{[g_{y,c}]_+^2}
{\sum_{c'}[g_{y,c'}]_+^2+\epsilon}.
\end{aligned}
\label{eq:shift_mass}
\end{equation}
As shown in Fig.~\ref{fig:further_analysis}(a), the top 10\%, 20\%, and 30\% of channels account for 82.80\%, 92.85\%, and 96.76\% of the positive corrected shift mass, with a mean Gini coefficient of 0.887. The number of significantly shifted channels also varies strongly across foreground classes. For example, \emph{sneezing} and \emph{pig} show broader affected subsets than \emph{blender} and \emph{crackling fire}. This class dependence supports input-dependent rather than globally fixed selection. This analysis is diagnostic rather than a causal identification result: IID and OOD routes contain different foreground recordings, and concentration refers specifically to the null-corrected statistic above. Neither test-set channel rankings nor this statistic supervise the selector; training uses source-class labels and observed background variation only.

\subsubsection{Shift and OOD Degradation}
We correlate episode-level representation shift with IID-to-OOD predictive degradation over 1,000 episodes. Figure~\ref{fig:further_analysis}(b) summarizes the mean correlations across three random seeds for accuracy, predicted probability, and classification margin. Across the three seeds, two shot settings, and three measures, all 18 Spearman correlations are positive, with $\rho$ between 0.268 and 0.455. Episodes with larger representation shift therefore tend to exhibit larger predictive degradation, establishing an association between the representation diagnosis and OOD performance.

\subsubsection{Selector Scores Reflect Functional Importance}
We rank the 640 ResNet12 channels by their selector scores and block the highest-scored, random, or lowest-scored 30\% while keeping the episodic classifier fixed. Figure~\ref{fig:further_analysis}(c) shows that blocking the highest-scored channels decreases accuracy by 3.033 percentage points on average, compared with 1.064 points for random blocking, while removing the lowest-scored channels produces almost no change. The learned scores therefore reflect clear differences in the functional contribution of representation channels.

Together, these analyses connect the empirical motivation in Sec.~\ref{sec:intro} with the behavior of the learned selector: background-induced changes are concentrated and class-dependent, larger representation shifts are associated with larger OOD degradation, and the selector assigns higher scores to channels with greater functional importance.

%% file: sections/05_conclusion.tex
\section{Conclusion}
\label{sec:conclusion}
We presented \method, a sample-conditioned fixed-budget representation selector for few-shot audio classification under class--background co-occurrence shift. The method keeps the source encoder frozen, learns per-example feature scores using source-class discrimination and cross-background contrastive supervision, and performs deterministic Top-$k$ selection at inference. Masks are per-example and support-only; no query label/statistic or test-time background label is used. On SpurAudio, \method\ achieves the best OOD accuracy in all four backbone/shot settings and improves the matched \fullrep\ control by 5.982, 4.902, 6.603, and 8.375 percentage points. Ablation and further analysis show that the gains depend on learned selection and cross-background supervision and are consistent with a representation structure in which background-induced changes are concentrated, class-dependent, and associated with predictive degradation.

%% file: references.bib
@misc{abuayoub2026spuraudio,
  title         = {{SpurAudio}: A Benchmark for Studying Shortcut Learning in Few-Shot Audio Classification},
  author        = {Abu Ayoub, Giries and Tukan, Morad and Mualem, Loay},
  howpublished  = {arXiv:2605.13672},
  year          = {2026},
  eprint        = {2605.13672},
  archiveprefix = {arXiv},
  doi           = {10.48550/arXiv.2605.13672},
}

@misc{heggan2022metaaudio,
  title         = {{MetaAudio}: A Few-Shot Audio Classification Benchmark},
  author        = {Heggan, Calum and Budgett, Sam and Hospedales, Timothy and Yaghoobi, Mehrdad},
  howpublished  = {arXiv:2204.02121},
  year          = {2022},
  eprint        = {2204.02121},
  archiveprefix = {arXiv},
  doi           = {10.48550/arXiv.2204.02121},
}

@inproceedings{vinyals2016matching,
  title     = {Matching Networks for One Shot Learning},
  author    = {Vinyals, Oriol and Blundell, Charles and Lillicrap, Timothy and Kavukcuoglu, Koray and Wierstra, Daan},
  booktitle = {Advances in Neural Information Processing Systems},
  volume    = {29},
  year      = {2016},
}

@inproceedings{snell2017prototypical,
  title         = {Prototypical Networks for Few-Shot Learning},
  author        = {Snell, Jake and Swersky, Kevin and Zemel, Richard S.},
  booktitle     = {Advances in Neural Information Processing Systems},
  volume        = {30},
  year          = {2017},
  eprint        = {1703.05175},
  archiveprefix = {arXiv},
}

@inproceedings{chen2019closer,
  title         = {A Closer Look at Few-Shot Classification},
  author        = {Chen, Wei-Yu and Liu, Yen-Cheng and Kira, Zsolt and Wang, Yu-Chiang Frank and Huang, Jia-Bin},
  booktitle     = {International Conference on Learning Representations},
  year          = {2019},
  eprint        = {1904.04232},
  archiveprefix = {arXiv},
}

@inproceedings{li2022taskspecific,
  title         = {Cross-Domain Few-Shot Learning With Task-Specific Adapters},
  author        = {Li, Wei-Hong and Liu, Xialei and Bilen, Hakan},
  booktitle     = {Proceedings of the IEEE/CVF Conference on Computer Vision and Pattern Recognition},
  pages         = {7161--7170},
  year          = {2022},
  eprint        = {2107.00358},
  archiveprefix = {arXiv},
}

@article{geirhos2020shortcut,
  title   = {Shortcut Learning in Deep Neural Networks},
  author  = {Geirhos, Robert and Jacobsen, J{\"o}rn-Henrik and Michaelis, Claudio and Zemel, Richard and Brendel, Wieland and Bethge, Matthias and Wichmann, Felix A.},
  journal = {Nature Machine Intelligence},
  volume  = {2},
  pages   = {665--673},
  year    = {2020},
  doi     = {10.1038/s42256-020-00257-z},
}

@inproceedings{dvornik2020sur,
  title         = {Selecting Relevant Features from a Multi-domain Representation for Few-Shot Classification},
  author        = {Dvornik, Nikita and Schmid, Cordelia and Mairal, Julien},
  booktitle     = {Computer Vision -- ECCV 2020},
  pages         = {769--786},
  year          = {2020},
  doi           = {10.1007/978-3-030-58607-2_45},
  eprint        = {2003.09338},
  archiveprefix = {arXiv},
}

@inproceedings{li2019categorytraversal,
  title         = {Finding Task-Relevant Features for Few-Shot Learning by Category Traversal},
  author        = {Li, Hongyang and Eigen, David and Dodge, Samuel and Zeiler, Matthew and Wang, Xiaogang},
  booktitle     = {Proceedings of the IEEE/CVF Conference on Computer Vision and Pattern Recognition},
  pages         = {1--10},
  year          = {2019},
  doi           = {10.1109/CVPR.2019.00009},
  eprint        = {1905.11116},
  archiveprefix = {arXiv},
}

@inproceedings{lee2022tdm,
  title         = {Task Discrepancy Maximization for Fine-Grained Few-Shot Classification},
  author        = {Lee, SuBeen and Moon, WonJun and Heo, Jae-Pil},
  booktitle     = {Proceedings of the IEEE/CVF Conference on Computer Vision and Pattern Recognition},
  pages         = {5331--5340},
  year          = {2022},
  eprint        = {2207.01376},
  archiveprefix = {arXiv},
}

@inproceedings{hu2018senet,
  title         = {Squeeze-and-Excitation Networks},
  author        = {Hu, Jie and Shen, Li and Sun, Gang},
  booktitle     = {Proceedings of the IEEE Conference on Computer Vision and Pattern Recognition},
  pages         = {7132--7141},
  year          = {2018},
  doi           = {10.1109/CVPR.2018.00745},
  eprint        = {1709.01507},
  archiveprefix = {arXiv},
}

@inproceedings{jang2017gumbelsoftmax,
  title         = {Categorical Reparameterization with {Gumbel-Softmax}},
  author        = {Jang, Eric and Gu, Shixiang and Poole, Ben},
  booktitle     = {International Conference on Learning Representations},
  year          = {2017},
  eprint        = {1611.01144},
  archiveprefix = {arXiv},
}

@inproceedings{kool2019gumbeltopk,
  title         = {Stochastic Beams and Where To Find Them: The {Gumbel-Top-k} Trick for Sampling Sequences Without Replacement},
  author        = {Kool, Wouter and Van Hoof, Herke and Welling, Max},
  booktitle     = {Proceedings of the 36th International Conference on Machine Learning},
  series        = {Proceedings of Machine Learning Research},
  volume        = {97},
  pages         = {3499--3508},
  publisher     = {PMLR},
  year          = {2019},
  eprint        = {1903.06059},
  archiveprefix = {arXiv},
}

@inproceedings{balin2019concrete,
  title         = {Concrete Autoencoders: Differentiable Feature Selection and Reconstruction},
  author        = {Bal{\i}n, Muhammed Fatih and Abid, Abubakar and Zou, James},
  booktitle     = {Proceedings of the 36th International Conference on Machine Learning},
  series        = {Proceedings of Machine Learning Research},
  volume        = {97},
  pages         = {444--453},
  publisher     = {PMLR},
  year          = {2019},
  eprint        = {1901.09346},
  archiveprefix = {arXiv},
}

@inproceedings{khosla2020supervised,
  title         = {Supervised Contrastive Learning},
  author        = {Khosla, Prannay and Teterwak, Piotr and Wang, Chen and Sarna, Aaron and Tian, Yonglong and Isola, Phillip and Maschinot, Aaron and Liu, Ce and Krishnan, Dilip},
  booktitle     = {Advances in Neural Information Processing Systems},
  volume        = {33},
  year          = {2020},
  eprint        = {2004.11362},
  archiveprefix = {arXiv},
}

@inproceedings{sgouropoulos2025prototypicalcontrastive,
  title         = {Prototypical Contrastive Learning for Improved Few-Shot Audio Classification},
  author        = {Sgouropoulos, Christos and Nikou, Christos and Vlachos, Stefanos and Theiou, Vasileios and Foukanelis, Christos and Giannakopoulos, Theodoros},
  booktitle     = {2025 IEEE 35th International Workshop on Machine Learning for Signal Processing},
  year          = {2025},
  doi           = {10.1109/MLSP62443.2025.11204215},
  eprint        = {2509.10074},
  archiveprefix = {arXiv},
}

@inproceedings{liu2026physics, title={From Physics to Representation: Audio Learning with Synthetic Pre-training via Procedural Generation}, author={Liu, Fengrui and Huang, Ruiyang and Zheng, Qijian and Wang, Yuanfang and Liu, Feng}, booktitle={Proceedings of the 2026 International Conference on Multimedia Retrieval}, pages={595--604}, year={2026} }

@article{liu2026prp, title={{PRP}: Procedural-to-Real Masked Pre-training for Transferable and Interpretable Audio Representations}, author={Liu, Fengrui and Huang, Ruiyang and Zheng, Qijian and Wang, Yuanfang and Liu, Feng}, journal={IEEE Transactions on Audio, Speech and Language Processing}, year={2026}, publisher={IEEE} }
